\documentclass[letterpaper, 10 pt, conference]{ieeeconf}  % Comment this line out if you need a4paper

\IEEEoverridecommandlockouts                              % This command is only needed if 
\usepackage[hyphens]{url}
\usepackage{hyperref}
\newenvironment{myitemize}{\begin{list}{$\bullet$}
{\setlength{\topsep}{1mm}
\setlength{\itemsep}{0.25mm}
\setlength{\parsep}{0.25mm}
\setlength{\itemindent}{0mm}
\setlength{\partopsep}{0mm}
\setlength{\labelwidth}{15mm}
\setlength{\leftmargin}{4mm}}}{\end{list}}

\usepackage{array}
\newcommand{\PreserveBackslash}[1]{\let\temp=\\#1\let\\=\temp}
\newcolumntype{C}[1]{>{\PreserveBackslash\centering}p{#1}}
\newcolumntype{R}[1]{>{\PreserveBackslash\raggedleft}p{#1}}
\newcolumntype{L}[1]{>{\PreserveBackslash\raggedright}p{#1}}

\usepackage{graphics} % for pdf, bitmapped graphics files
\usepackage{epsfig} % for postscript graphics files
\usepackage{mathptmx} % assumes new font selection scheme installed
\usepackage{times} % assumes new font selection scheme installed
\usepackage{amsmath} % assumes amsmath package installed
\usepackage{amssymb}  % assumes amsmath package installed
\usepackage{multirow}
\usepackage{bm}
\usepackage{graphicx}
\usepackage{xcolor}
\usepackage{diagbox}

\usepackage[ruled,linesnumbered]{algorithm2e}
\SetKwRepeat{Do}{do}{while}

\title{\LARGE \bf Interactive Trajectory Planner for Mandatory Lane Changing\\in Dense Non-Cooperative Traffic}

\author{Xiangguo Liu$^{1}$, Jianxing Chen$^{2}$, Shan Li$^{2}$, Yajia Zhang$^{2}$, Hongtao Yu$^{2}$\\Fuqiang Huang$^{2}$, Jiechao Liu$^{2}$, Chao Wang$^{2}$, Liyun Li$^{2}$, Qi Zhu$^{1}$% <-this % stops a space
\thanks{$^{1}$Xiangguo Liu and Qi Zhu are with the Department of Electrical and Computer Engineering, Northwestern University, Evanston, IL 60201, USA.
        {\tt\small xg.liu@u.northwestern.edu, qzhu@northwestern.edu.}}%
\thanks{$^{2}$Jianxing Chen, Shan Li, Yajia Zhang, Hongtao Yu, Fuqiang Huang, Jiechao Liu, Chao Wang and Liyun Li are with the Autonomous Driving Center, XMotors.ai, Mountain View, CA 94043, USA.
        }%
}

\begin{document}

\maketitle
\thispagestyle{empty}
\pagestyle{empty}

%%%%%%%%%%%%%%%%%%%%%%%%%%%%%%%%%%%%%%%%%%%%%%%%%%%%%%%%%%%%%%%%%%%%%%%%%%%%%%%%
\begin{abstract}
When the traffic stream is extremely congested and surrounding vehicles are not cooperative, the mandatory lane changing can be significantly difficult. In this work, we propose an interactive trajectory planner, which will firstly attempt to change lanes as long as safety is ensured. Based on receding horizon planning, the ego vehicle can abort or continue changing lanes according to surrounding vehicles' reactions. We demonstrate the performance of our planner in extensive simulations with eight surrounding vehicles, initial velocity ranging from 0.5 to 5 meters per second, and bumper to bumper gap ranging from 4 to 10 meters. The ego vehicle with our planner can change lanes safely and smoothly. The computation time of the planner at every step is within 10 milliseconds in most cases on a laptop with 1.8GHz Intel Core i7-10610U. 
\end{abstract}

%%%%%%%%%%%%%%%%%%%%%%%%%%%%%%%%%%%%%%%%%%%%%%%%%%%%%%%%%%%%%%%%%%%%%%%%%%%%%%%%
\section{Introduction}
Trajectory planning for autonomous vehicles has been an active research topic for many years~\cite{liu2022markov}. Nowadays it is even been developed and adopted in production vehicles by some automotive companies, such as Tesla and Xmotors.ai. However, there are still some challenging tasks, which limit its application scenarios. Mandatory lane changing in dense traffic is one of the most challenging tasks, which is even difficult for human drivers.

Different from car following, lane changing is widely recognized to be challenging for several reasons: (a) It requires coordination of longitudinal and lateral motion planning. It is usually not a feasible solution to first complete lateral movement and then adjust velocity longitudinally, or vice versa, especially in congested scenarios. (b) It takes the responsibility to avoid collision~\cite{zhang2022comfort} with vehicles in both the original lane and the desired lane. (c) In dense traffic, there is usually not enough gap in the desired lane to complete lane changing. The ego vehicle has to interact and compete with surrounding vehicles to change lanes~\cite{liu2023safety}. Furthermore, it could be that surrounding vehicles do not cooperate and ignore the turning signals. (d) It needs real-time computation performance in a dynamic environment, given more constraints.

\begin{figure}[tbp]
    \centering\includegraphics[scale=0.45]{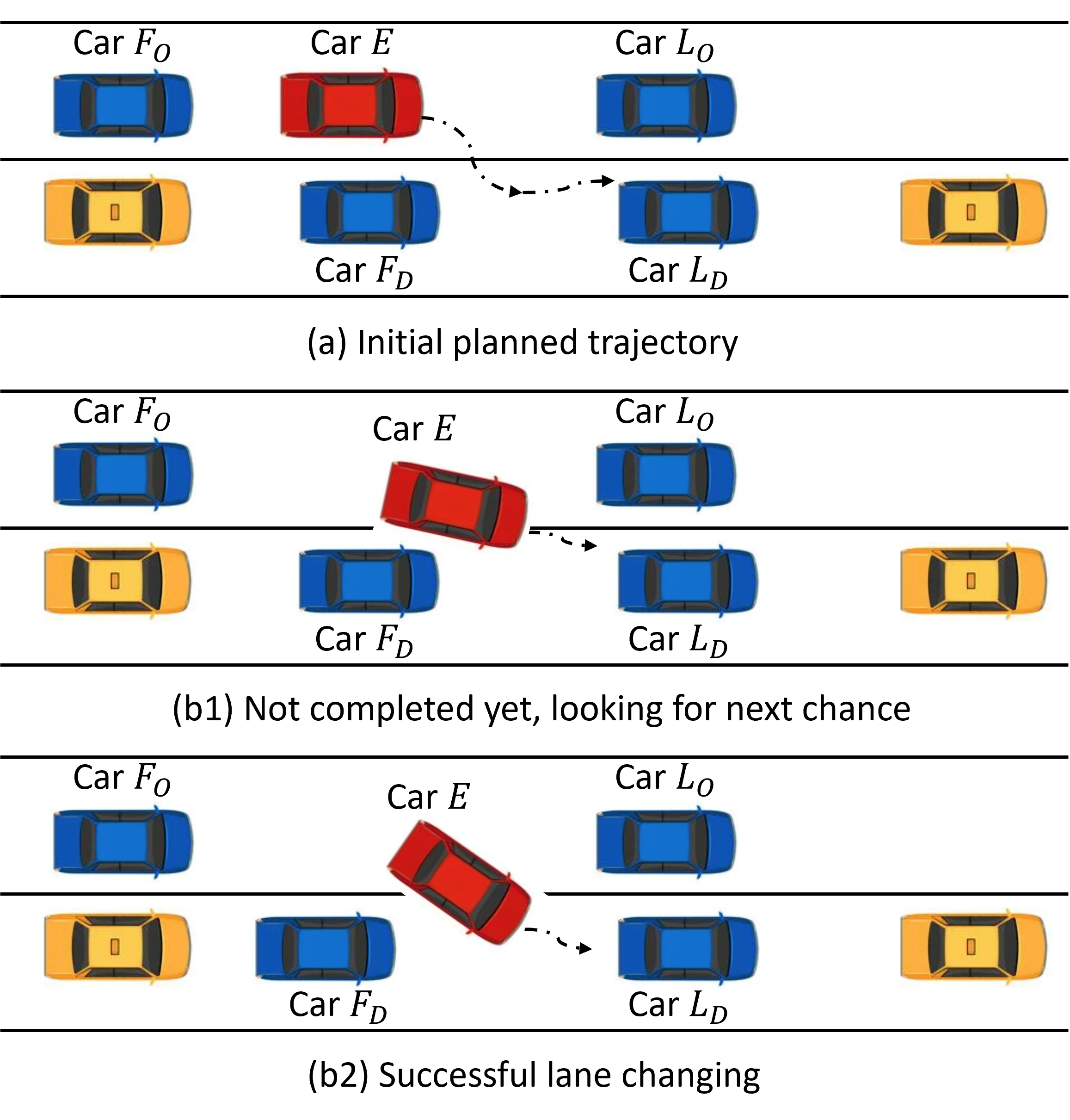}
    \caption{The ego vehicle $E$ intends to change lanes and is surrounded by vehicle $F_O$, $L_O$, $F_D$ and $L_D$. $F$ and $L$ denote following and leading vehicle, respectively. The subscript $O$ and $D$ represent the original and the desired lane, respectively. In subplot (a), the ego vehicle $E$ cannot find a feasible trajectory to complete lane changing in dense traffic. It makes an attempt first with slight lateral movements. In some cases as shown in subplot (b1), the ego vehicle $E$ still cannot change lanes successfully after the attempt, because vehicle $F_D$ does not cooperate and the gap remains small. In other cases as shown in subplot (b2), the ego vehicle $E$ occupies a favorable position over vehicle $F_D$, and forces it to decelerate, otherwise there might be collisions. The ego vehicle $E$ changes lane successfully in a larger gap.\label{fig:scenarios}}
\end{figure}

This work focuses on those dense low-speed scenarios that surrounding vehicles ignore other vehicles' turning signals. The ego vehicle has to interact to force others to cooperate, otherwise there is no chance for lane changing. To overcome these difficulties, we propose the attempt-first interactive lane changing planner. The idea is illustrated in figure~\ref{fig:scenarios}. The ego vehicle $E$ intends to change lanes and needs to avoid collision with $L_O$, $F_O$, $L_D$ and $F_D$, which are the leading and following vehicles in the original lane and the desired lane, respectively. Initially, ego vehicle $E$ cannot find a large gap and safely change lanes. But it can make a slight lateral movement and attempt to change lanes as shown in the subplot (a). While taking the attempt, the ego vehicle is able to abort lane changing and move back to the original lane safely. At the same time, it can interact and put pressure on vehicle $F_D$, which may decelerate to enlarge the gap. In this work, we assume a dense and non-cooperative traffic environment, and surrounding vehicles will not decelerate unless it may result in accidents otherwise. In some cases, after the attempt, the ego vehicle $E$ still could not complete lane changing if the gap remains small, as shown in subplot (b1). It will keep looking for the next chance to change lanes, as in our receding horizon planning framework. In other cases as shown in subplot (b2), the ego vehicle $E$ can occupy a favorable position with the attempt, then force vehicle $F_D$ to decelerate, and change lanes successfully at the end.

This planner is developed based on a recently proposed optimization algorithm, Constrained Iterative Linear Quadratic Regulator (CILQR), which can provide real-time planning results. It transforms original optimization problem with non-convex inequality constraints to a newly formulated problem with a quadratic cost function and no constraints, thus improving computation efficiency. However, those soft constraints can be violated in this way. We conduct safety analysis for the planned trajectory, and if necessary, we adjust the parameters in the cost function (including the desired path) for a safer and smoother trajectory. 

In summary, our work makes the following contributions:
\begin{myitemize}
\item We incorporate vehicular interactions into trajectory planner design for lane changing in dense non-cooperative traffic. The ego vehicle can make a lane changing attempt first, and then abort or continue the behavior as a safe and fast response to surrounding vehicles' reactions.

\item We develop a receding horizon planning framework, in which the ego vehicle can keep looking for the chance to change lanes and react to surrounding vehicles quickly, thus improving safety and lane changing success rate.

\item We implement the CILQR algorithm in C++ to solve the non-convex constrained optimization problem in real time and conduct extensive experiments to demonstrate our planner's effectiveness for lane changing in dense traffic.
\end{myitemize}

The rest of this paper is organized as follows. In Section~\ref{sec:related_work}, we review related works for trajectory planning and vehicular interaction. In Section~\ref{sec:Methodology}, we present our planner design. We then present the experimental setting and analyze the simulation results in Section~\ref{sec:experiments}. We summarize the takeaways in Section~\ref{sec:conclusion}.

\section{Related Works}\label{sec:related_work}
Our work is related to a rich literature on trajectory planning, and more specifically, lane changing. Some of those also consider vehicular interactions, which we will discuss separately in Section~\ref{sec:interaction_in_dense} in detail.

\subsection{Trajectory Planning and Lane Changing}

A large number of trajectory planning techniques are proposed and developed to improve safety and performance~\cite{chen2023mixed} of the transportation system. Those can be roughly classified into several categories: search-based, sampling-based, optimization-based~\cite{xiao2022robotic,wu2020amphibious} and learning-based methods. Those methods have their own strength in different application scenarios. \cite{katrakazas2015real,gonzalez2015review} review these methods thoroughly. 

{Both search-based and sampling-based methods are developed in a discretized space (either state space or action space), thus} the computation efficiency will be significantly influenced if we increase the discretization precision. Besides this point, search-based methods are limited in spatiotemporal planning while {sampling-based methods may lead to jerky paths.} Learning-based methods greatly improve the performance of complex systems, while saving effort and time in modeling system dynamics. However, safety is hard to be verified for these methods, especially in those near-accident scenarios~\cite{cao2020reinforcement}.

Optimization-based methods, e.g., Model Predictive Control (MPC), can handle static and dynamic obstacles as constraints and output a smooth trajectory intrinsically. However, these constraints from obstacles and road boundaries are usually non-convex. It considerably increases the computation time for the optimization problem, which is hard to be in real time~\cite{liu2021trajectory}. Recent proposed algorithms, Differential Dynamic Programming (DDP)~\cite{tassa2014control}, Iterative Linear Quadratic Regulator (ILQR)~\cite{van2014iterated}, and most recently, CILQR~\cite{chen2017constrained,chen2019autonomous}, can solve real-time optimization with non-convex constraints within 200 milliseconds. We implement and develop the CILQR algorithm in C++ for our work and further reduce the time to be within 10 milliseconds.

As one of the most challenging driving maneuvers and one of the main causes of congestion and collisions, lane changing is comprehensively reviewed in~\cite{rahman2013review,zheng2014recent}. \cite{dang2014lane,luo2016dynamic} propose a dynamic lane changing planner, which can update its reference trajectory periodically. It can plan a trajectory back to the original lane to eliminate collision. \cite{yang2018dynamic} has the similar idea with \cite{luo2016dynamic}. But they all assume that leading and following vehicles in the target lane will remain their velocities when the ego vehicle changes lane, which does not consider interactions between vehicles and environmental disturbance. Some learning-based lane changing planners~\cite{shi2019driving,alizadeh2019automated,ye2020automated} are developed to decide the starting time and trajectory, but without safety guarantee. 

Next we will discuss those works that consider interaction for lane changing in dense traffic, and compare those with our planner.

%\cite{dang2014lane} same authors with \cite{luo2016dynamic}, published in ITSC.

%\cite{yang2018dynamic} similar to \cite{luo2016dynamic}, it can plan a trajectory back to the original lane to eliminate collision. \cite{yang2018dynamic,luo2016dynamic} assume that leading and following vehicles in the target lane will remain their velocities when the ego vehicle changes lane, which does not consider the driving fluctuations and interactions between vehicles.

%\cite{shi2019driving} developed hierarchical RL planners to address when and how to change lane, respectively. 
%\cite{alizadeh2019automated} RL for lane changing.
%\cite{ye2020automated} RL for lane changing, considering the action of aborting lane changing and return back to original lane.
%All these three have no safety guarantee.

%\cite{sivaraman2014dynamic} Dynamic Probabilistic Drivability Maps for Lane Change and Merge Driver Assistance, leverage dynamic programming to compute acceleration and timing to safely merge or change lanes.

\subsection{Interaction in Dense Traffic}\label{sec:interaction_in_dense}

Game theory based models~\cite{ali2019game,yu2018human,9304804} are common methods to model interactions between vehicles, which are reviewed comprehensively in~\cite{talebpour2015modeling}. \cite{ladino2020dynamic} leverages game theory to model cooperative lane changing scenarios, in which vehicles aim to minimize the joint cost function. %\cite{awal2015efficient} assumes vehicles are all autonomous and cooperative. This complete cooperation assumption may not be realistic nowadays. 
\cite{zhang2019game} simplifies the lateral movement of the subject vehicle, which is a binary variable that represents lanes. It is assumed that vehicles interact with turning signals, which does not fit in the congested scenarios well~\cite{kauffmann2018makes}. \cite{talebpour2015modeling} proposes a game theory based lane changing model for connected vehicles, and calibrates different parameters in the utility functions for mandatory and discretionary lane changing scenarios. \cite{ali2019game} also calibrates parameters from traffic datasets, which may not perform well for one specific driver or scenario.

\cite{yu2018human} defines a parameter called aggressiveness to represent driver's personality, estimates it in real time and incorporates it in the utility function. Different extent of aggressiveness represents a different preference for travel time, headway and etc. It models vehicle's interaction in lane changing process in a Stackelberg game. However, the extent of aggressiveness may change in the interaction process, and it is hard to accurately estimate the value for human drivers. An interesting question from the analysis of the model is that, should the ego vehicle just give up lane changing if surrounding vehicles behave very aggressively? Or can the ego vehicle always be more aggressive than others to improve lane changing success rate? Our planner design demonstrates that in extremely dense traffic, even though surrounding vehicles perform aggressively, the ego vehicle can still make the lane changing attempt, as long as it is safe. It may be that the following vehicle decelerates a little bit later, or the leading vehicle accelerates, then the lane changing can be completed. 

Machine learning methods have natural strengths in interaction modeling and prediction~\cite{jiao2022semi,jiao2022tae,9991055}. \cite{hou2013modeling} makes discrete behavior decisions for mandatory lane changing based on Bayes classifier and decision trees. %\cite{burger2020interaction} obtains a probability distribution of different intentions of surrounding vehicles by Bayesian estimation, then plans trajectories considering uncertain interaction with surrounding vehicles. 
\cite{hubmann2018belief} leverages Partially Observable Markov Decision Process (POMDP) to include surrounding vehicles in the state space, thus planning interactive behavior. Similarly, \cite{bouton2019cooperation} leverages POMDP to model the level of cooperation of other drivers, and incorporate this belief into reinforcement learning for a higher merging success rate. \cite{bae2020cooperation} leverages RNN to model interaction between vehicles, then the predicted results will be used in safety constraints of the MPC planner. \cite{chen2019attention} develops hierarchical deep reinforcement learning for lane change behaviors in autonomous driving. \cite{bouton2020reinforcement} acquires a strategic level k planner for merging in dense traffic by reinforcement learning and iterative reasoning. But these learning-based methods are hard and expensive to assure safety~\cite{liu2022physics,zhu2021safety, wang2020energy, wang2022design,liu2023connectivity,chang2023safety}.

These methods have two common limitations. Firstly, these methods assume the ego vehicle can only choose one among several optional actions, e.g., change lanes, go back to original lane and do nothing. However, it is common that a driver makes a lightly lateral movement but hesitates to complete lane changing due to safety concerns. These ambiguous behaviors are rarely considered. Secondly, with a predefined model for surrounding vehicles, it may results in safety issues if there are model mismatches. In fact, modeling human drivers is complicated due to the irrational and dynamical characteristics. 

Some works attempt to ensure safety in the worst case and assuming surrounding vehicles may behave unpredictably. \cite{naumann2019provably} computes safety distance for the ego vehicle assuming it brakes to prevent collision whenever necessary. The ego vehicle initiates lane changing if the safety distance is satisfied, thus safety is assured. However the evasion behavior does not consider steering, the safety distance in fact is hard to be satisfied in dense traffic. \cite{shalev2017formal,pek2017verifying} have similar limitations.

Our work is closely related to~\cite{chandru2017safe}. It analyzes the critical distance from surrounding vehicles, thus ensuring safety of the ego vehicle by staying out of the critical region. It considers the evasion behavior of braking and steering. However, it first decides the target state at the end of planning horizon, then plans the trajectory with this constraint. While in our planner design, the ego vehicle has no final state constraint and is encouraged to move laterally towards the desired path quickly. In this way, the ego vehicle can put pressure on surrounding vehicles as soon as possible. In Section~\ref{sec:experiments}, we demonstrate that an inappropriate target state (or desired path) may result in a trajectory with a large yaw angle, or even no feasible solution. Besides that, we consider a more realistic kinematic model, and our planner's real time performance is much better.

\section{Methodology}
\label{sec:Methodology}

\begin{figure*}[tbp]
    \centering\includegraphics[scale=0.43]{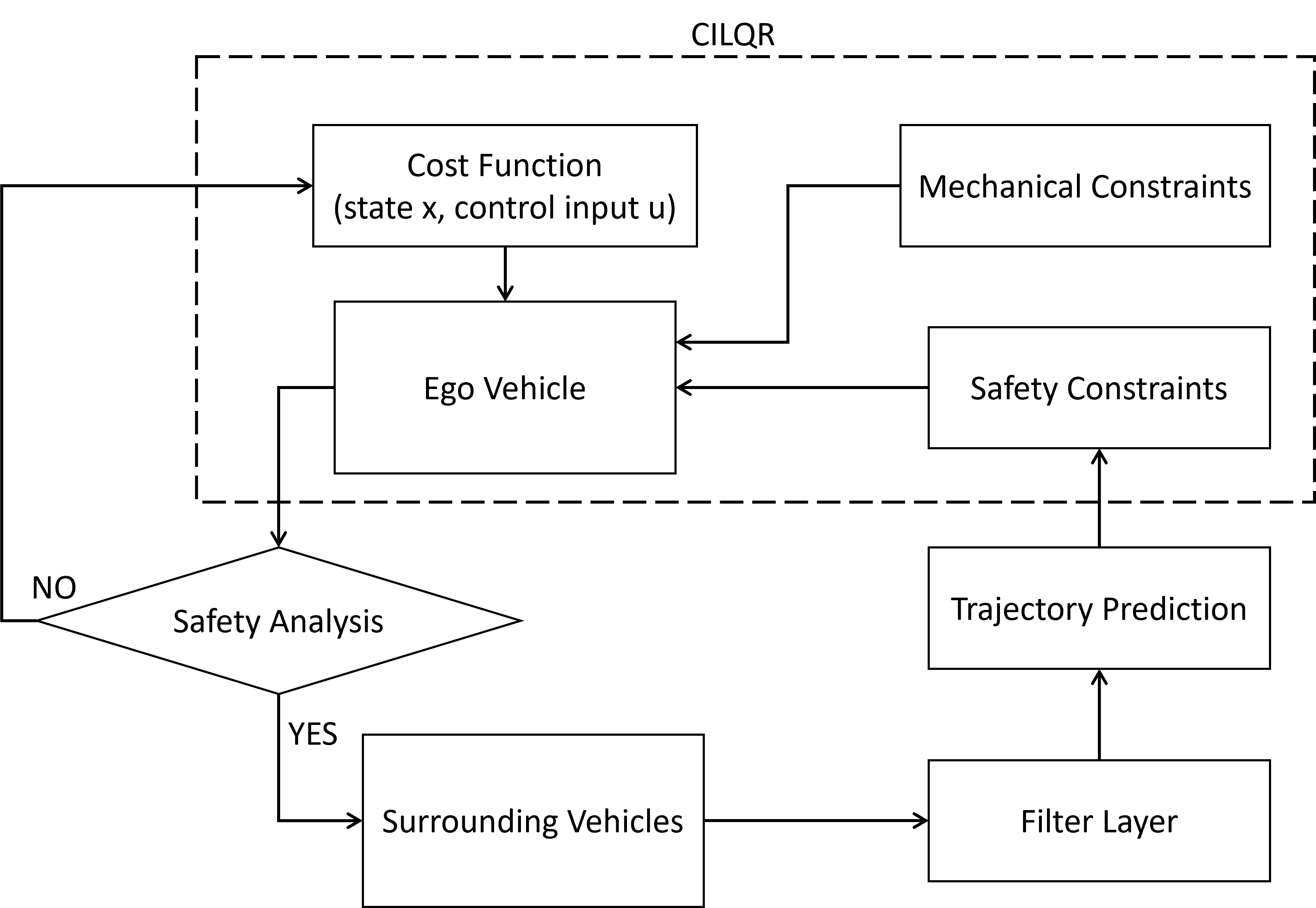}
    \caption{Interactive planner design for lane changing in dense traffic. Surrounding vehicles' motion statuses are processed by the filter layer first, which extracts all directly adjacent vehicles. Then these surrounding vehicles' trajectories are predicted and transformed into safety constraints for the ego vehicle's motion. With all these safety constraints and the mechanical constraints of the ego vehicle itself, the ego vehicle's trajectory is planned to minimize our designed cost function by CILQR algorithm. Then we analyze the planned trajectory and execute it if safety is assured, otherwise adjust cost function to update the planned trajectory. It is noted that modifying the desired path in the cost function, e.g., aborting changing lanes and staying in the original lane, should always provide a safe trajectory.\label{fig:overview}}
\end{figure*}

In this work, we focus on lane changing in dense and non-cooperative traffic scenarios, in which traditional rule-based planners may not be able to find a gap to change lanes. We propose the planner design that leverages vehicular interactions to gradually enlarger the gap and finally change lanes successfully. The design is shown in figure~\ref{fig:overview}. In congested and non-cooperative traffic scenarios, surrounding vehicles all follow their leading vehicles closely, but still interact to prevent collisions. During the interaction process, surrounding vehicles' motion statuses are processed by the filter layer first, which extracts all possibly directly adjacent vehicles in the next few seconds. Then these surrounding vehicles' trajectories are predicted and transformed into safety constraints for the ego vehicle's motion. With all these safety constraints and the mechanical constraints of the ego vehicle itself, the ego vehicle's trajectory is planned to minimize our designed cost function. In this work, we develop and implement the CILQR algorithm in C++ to acquire planning results of this non-convex optimization problem in real time. Since these non-convex constraints are transformed and added to the cost function in CILQR, we analyze the planned trajectory and execute it if safety is assured, otherwise adjust cost function to update the planned trajectory. It is noted that modifying the desired path in the cost function, e.g., aborting changing lanes and staying in the original lane, should always provide a feasible trajectory.

The ego vehicle is modeled by the kinematic bicycle model as in~\cite{polack2017kinematic,chen2017constrained}, and the state is $x=[p_x, p_y, v, \theta]^\mathrm{T}$. $p_x$ and $p_y$ are longitudinal and lateral positions, respectively. $v$ is the velocity and $\theta$ is yaw angle. The control input is $u=[\dot v, \dot \theta]^\mathrm{T}=[a, \frac{v\cdot tan(\delta)}{L}]^\mathrm{T}$, where $a$ is acceleration, $\delta$ is the steering angle of front wheels and $L$ is the wheelbase. The system dynamics can be formulated as
\begin{gather}\label{eq:dynamics}
 \dot x = \begin{bmatrix} \dot p_x \\ \dot p_y \\ \dot v \\ \dot \theta \end{bmatrix}
 =
  \begin{bmatrix}
  v \cdot cos(\theta) \\ v \cdot sin(\theta) \\ 0 \\ 0
   \end{bmatrix}
   +
   \begin{bmatrix}
  0 && 0 \\ 0 && 0 \\ 1 && 0 \\ 0 && 1
   \end{bmatrix}
   \times
   \begin{bmatrix}
  \dot v \\ \dot \theta
   \end{bmatrix}
\end{gather}

In the cost function $J(x, u)$, we consider penalty for large acceleration $a$ and yaw rate $\dot \theta$, penalty for low velocity $v$ and penalty for large mismatch between the planned path $\{(p_x, p_y)_i\}$ and the desired path
$\{(d_x, d_y)_i\}$. $(p_x, p_y)_i$ is the planned position of the ego vehicle at time $t=t_i$, and $(d_x, d_y)_i$ is the closest position to $(p_x, p_y)_i$ on the desired path. These cost terms are all quadratic for efficient computation as in~\cite{chen2017constrained}. It is noted that in our work, the coefficients for these cost terms and especially the desired path $\{(d_x, d_y)_i\}$ are all effective parameters to tune for a safe planned trajectory.

The mechanical constraints include lower and upper bounds for acceleration $a$ and yaw rate $\dot \theta$,
\begin{equation} \label{eq:mech_constraints}
\begin{cases}
%\begin{aligned}
a_{min} \leq a \leq a_{max}
\\
\dot \theta_{min} \leq \dot \theta \leq \dot \theta_{max}
%\end{aligned}
\end{cases}
\end{equation}
where $\dot \theta_{min}$ and $\dot \theta_{max}$ are determined by the minimum and maximum steering angle, $\dot \theta_{min}=\frac{v\cdot tan(\delta_{min})}{L}$ and $\dot \theta_{max}=\frac{v\cdot tan(\delta_{max})}{L}$.

The safety constraints prevent the ego vehicle from collisions with any surrounding vehicle. Given the states of ego vehicle $x_e$ and any surrounding vehicle $x_{s,k}$, $\forall k \in \{0, 1, 2, \cdots\}$, as well as the vehicle dimensions, the minimum distance between vehicles $g(x_e, x_{s,k})$ should be larger than the predefined safety margin $s_{min}$. Here we model surrounding vehicles as ellipses, in which the long axis $a_{s,k}$ and short axis $b_{s,k}$ are derived from the vehicle dimension as well as its velocity. We model ego vehicle as two circles, which are at $(p_{x,e,1}, p_{y,e,1})$ and $(p_{x,e,2}, p_{y,e,2})$, the center of the front-axis and rear-axis, respectively. The radius of the circle, $r_e$, is mainly based on the car's dimension. 
\begin{equation} \label{eq:safety_constraints}
\begin{cases}
%\begin{aligned}

%\begin{gather}
 \begin{bmatrix} p^{s,k}_{x,e,\eta} \\  p^{s,k}_{y,e,\eta} \end{bmatrix}
 =\begin{bmatrix}
  cos(\theta_{s,k}) && sin(\theta_{s,k}) \\ -sin(\theta_{s,k}) && cos(\theta_{s,k})
   \end{bmatrix}
   \times
   \begin{bmatrix}
  -p_{x,s,k}+p_{x,e,\eta} \\ -p_{y,s,k}+p_{y,e,\eta}
   \end{bmatrix}
%\end{gather}

\\
\frac{(p^{s,k}_{x,e,\eta})^2}{(a_{s,k}+r_e+s_{min})^2} + \frac{(p^{s,k}_{y,e,\eta})^2}{(b_{s,k}+r_e+s_{min})^2}\geq 1 
\\ 
\forall k \in \{0, 1, 2, \cdots\}, \eta \in \{1,2\}
%\end{aligned}
\end{cases}
\end{equation}
where $(p_{x,e,\eta}, p_{y,e,\eta})$ is longitudinal and lateral positions of the center of the front-axis and rear-axis, respectively. $\eta=1$ or $2$ correspond to the front-axis and the rear-axis, respectively. $p_{x,s,k}$, $p_{y,s,k}$ and $\theta_{s,k}$ are longitudinal position, lateral position and yaw angle of the surrounding vehicle $k$. $p^{s,k}_{x,e,\eta}$ and $p^{s,k}_{y,e,\eta}$ are positions of ego vehicle in the coordinate system of the surrounding vehicle $k$. %See ~\cite{chen2017constrained} for more details.

In extremely congested traffic scenarios, we need to pay more attention to the long axis $a_{s,k}$ of the ellipse representing surrounding vehicle $k$. For a large $a_{s,k}$, it is probably infeasible to change lanes due to violation of these safety constraints. While $a_{s,k}$ is small, it is likely that we underapproximate the reachable region of the surrounding vehicle, thus resulting in collisions 
in real life. Moreover, as surrounding vehicles are not static, their states $x_{s,k}$ need to be predicted over the planning horizon $T$. In fact, the prediction accuracy and the long axis are closely related. If we are able to predict surrounding vehicles' trajectories more accurately, we can set a smaller $a_{s,k}$, thus improving lane changing success rate. In this work, we predict the trajectories of surrounding vehicles as they will keep their latest velocities over the next planning horizon $T$. As demonstrated in section~\ref{sec:experiments}, in congested and low-velocity scenarios, the assumption works well with the receding horizon planning method. For future work, $a_{s,k}$ can also be adaptively adjusted based on surrounding vehicles' reactions, as they may be more aggressive or cautious at different times. 

With these constraints and the cost function, the planning problem can be formulated as below,
\begin{equation} \label{eq:planning}
%\begin{cases}
\begin{aligned}
\min_{u_0, u_1, \cdots, u_{N-1}} & J(x,u)=
\frac{1}{2}x_N^\mathrm{T}Q_Nx_N + x_N^\mathrm{T}q_N+p_N+
\\
&\sum_{i=0}^{N-1} (\frac{1}{2}x_i^\mathrm{T}Q_ix_i + x_i^\mathrm{T}q_i + \frac{1}{2}u_i^\mathrm{T}R_iu_i + u_i^\mathrm{T}r_i + p_i)
\\
s.t.\quad & x_{i+1}=f(x_i, u_i)=x_i+\dot x_i \cdot \delta t,\quad i=0, \cdots, N-1
\\
& a_{min} \leq a_i \leq a_{max},\quad i=0, \cdots, N-1
\\
& \dot \theta_{min} \leq \dot \theta_i \leq \dot \theta_{max},\quad i=0, \cdots, N-1
\\
& g(x_i, x_{s, k, i}) \geq s_{min}, \quad \forall k =0, 1, 2,\cdots, \forall i=0, \cdots, N
\end{aligned}
%\end{cases}
\end{equation}
where $x_i$ and $u_i$ is the state of ego vehicle and control input at the $i_{th}$ step, respectively. The planning horizon is $T=N \cdot \delta t$ and $\delta t$ is time stepsize. The state at next step $x_{i+1}$ is determined by control input $u_i$ and the state $x_i$ at time $t_i=i \cdot \delta t$. $\dot x_{i}$ is computed from equation~\ref{eq:dynamics}. $Q_i$, $q_i$, $R_i$, $r_i$ and $p_i$ are all coefficients for different penalty terms.

In a receding horizon planning framework, the above optimization problem is solved every $\lambda \cdot \delta t$ time. With $\lambda < N$, we can acquire $N$ control inputs from the solution, but will only apply the first $\lambda$ control inputs on the system. It can improve system safety and performance in a dynamic traffic environment.

However, the computation efficiency is usually a bottleneck of these non-convex optimization problems. In this work, we implement CILQR algorithm in C++ to solve the problem efficiently. It costs less than 10 milliseconds when there are 8 surrounding vehicles in most cases on a laptop with 1.8GHz Intel Core i7-10610U. This algorithm linearizes these constraints, derives barrier functions from them, quadratizes the barrier functions and adds these terms to the cost function. For the new optimization problem with only quadratic cost functions and without constraints, it can be solved in real time with existing methods. More details about CILQR algorithm can be found in ~\cite{chen2017constrained,chen2019autonomous}. 

This algorithm improves computation efficiency greatly, but cannot tell whether these constraints are violated or not. The original optimization problem has hard constraints, while the modified problem has only soft constraints. We conduct safety analysis for the solution acquired by the algorithm. If those constraints are not satisfied, in congested and low-velocity scenarios, slowing down and staying on current lateral position is at least a safe backup solution.   

Let us assume the worse case, if not the worst, surrounding vehicles ignore the turning signal and lane changing intention of the ego vehicle, and they do not cooperate. But at least, surrounding vehicles will decelerate if it is possible to collide otherwise, as formulated below.
\begin{equation} \label{eq:surrounding}
\dot v_{s,k}=
\begin{cases}
%\begin{aligned}
\max (a_{min}, -\frac{v_{s,k}}{\delta t}) \qquad \text{if }\exists \text{ vehicle } \alpha \text{ such that }
\\ \qquad 0 \leq p_{x,\alpha}-p_{x,s,k} \leq s_k \text{, } |p_{y,\alpha}-p_{y,s,k}| \leq r_{\alpha} + \frac{w_{s,k}}{2}
\\
\min (a_{max}, \frac{v_{max}-v_{s,k}}{\delta t}) \quad \text{otherwise }
%\end{aligned}
\end{cases}
\end{equation}
where $v_{s,k}$ and $\dot v_{s,k}$ are velocity and acceleration of surrounding vehicle $k$, respectively. Vehicle $\alpha$ can be the ego vehicle or any other surrounding vehicle. $s_k$ is the distance threshold for vehicle $k$ to avoid collision. $w_{s,k}$ is the width of vehicle $k$. Similarly, we model vehicle $\alpha$ as a circle with radius of $r_{\alpha}$.

Within our safe planning framework, we set the desired path in the target lane initially. It is probably that the solution that we acquired cannot complete the lane changing process in dense traffic. As long as it is safe and does not violate those constraints, the ego vehicle can execute the first $\lambda$ control inputs every time. Then in the dynamic traffic environment, the ego vehicle keeps finding opportunities to occupy a favorable position, thus putting pressure on surrounding vehicles. Once they decelerate and the gap is large enough, the ego vehicle can complete lane changing successfully.

\section{Experiments and Analysis}
\label{sec:experiments}
\begin{figure}[tbp]
    \centering\includegraphics[scale=0.5]{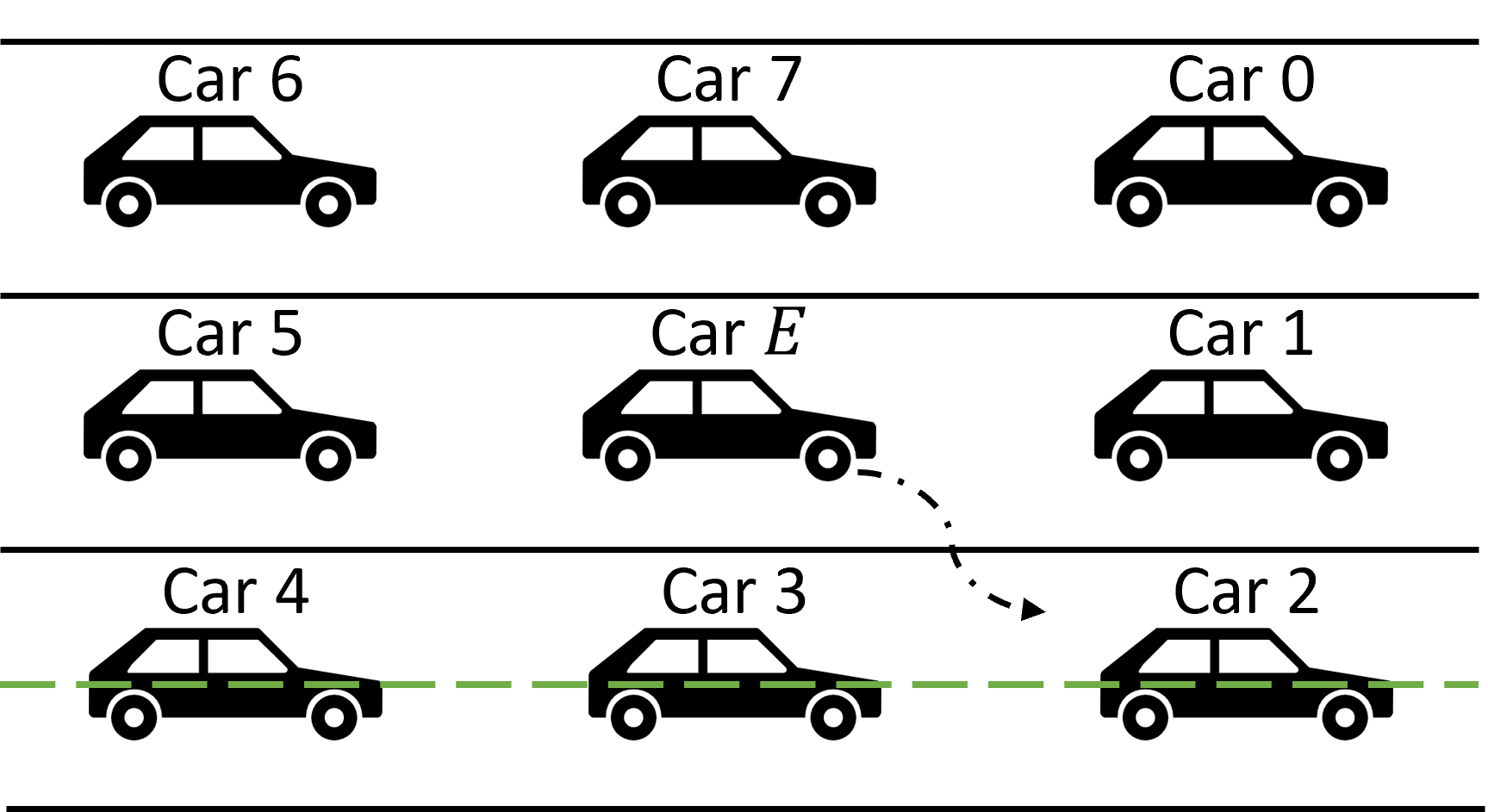}
    \caption{Eight surrounding vehicles with index of 0 to 7 are around the ego vehicle $E$. The ego vehicle $E$ intends to change lanes and the desired path is marked with green dash line. Initially all vehicles have the same speed $v_0$ and keep the same bumper to bumper distance $d_0$ with leading vehicles.\label{fig:simulation}}
\end{figure}

%\subsection{Simulation Design}
To simulate lane changing in dense traffic, we have eight surrounding vehicles (with index 0-7) around the ego vehicle $E$ as shown in figure~\ref{fig:simulation}. The desired lane is marked with green dash line. We assume all vehicles have the same speed $v_0$ initially. The initial bumper to bumper distances $d_0$ between any two longitudinally adjacent vehicles are the same. The planning horizon is $T=4$ seconds and time stepsize is $\delta t=0.1$ seconds. The planned trajectory is updated every $\delta t$ time with the receding horizon planning method.

\begin{figure*}[tbp]
    \centering\includegraphics[scale=0.34]{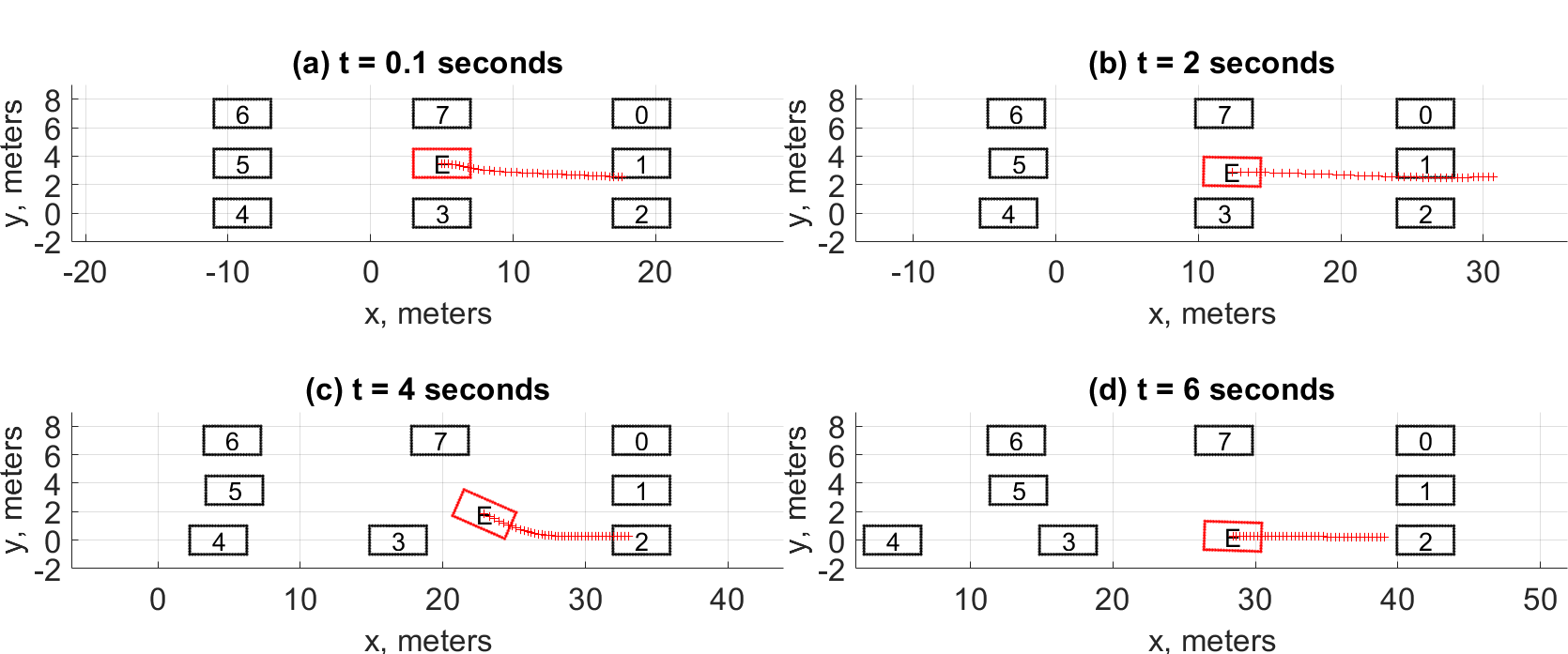}
    \caption{The x and y axes show the longitudinal and lateral positions of vehicles. Four subplots show the positions at different times. Red rectangle represents the ego vehicle, and black rectangles are surrounding vehicles. The planned trajectories of ego vehicle are plotted in red line. It corresponds to the scenario that initial velocity is $v_0=2$ m/s for all vehicles, and bumper to bumper distance $d_0=10$ m.\label{fig:vel2gap10}}
\end{figure*}

\begin{figure*}[tbp]
    \centering\includegraphics[scale=0.34]{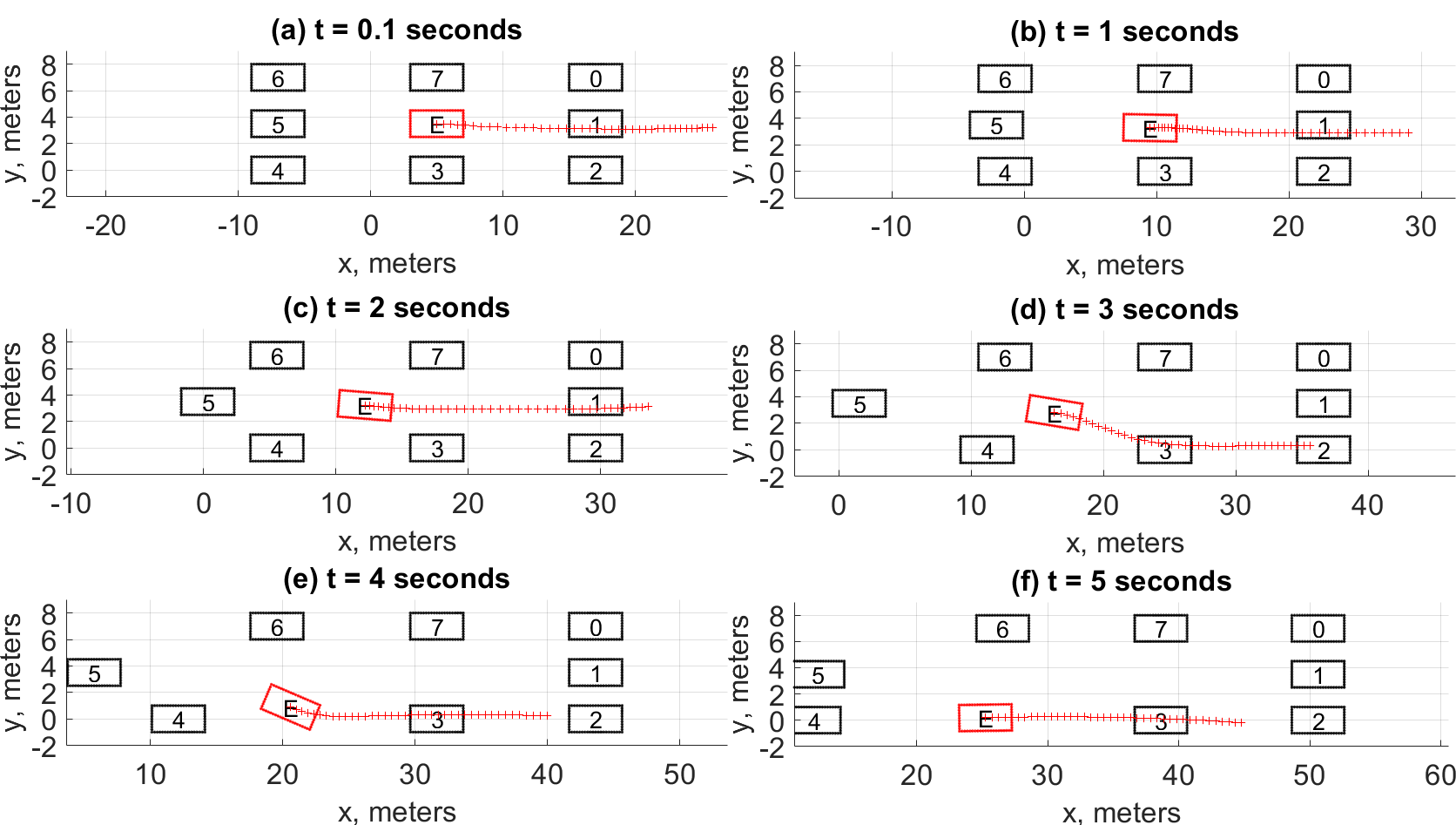}
    \caption{The x and y axes show the longitudinal and lateral positions of vehicles. Six subplots show the positions at different times. Red rectangle represents the ego vehicle, and black rectangles are surrounding vehicles. The planned trajectories of ego vehicle are plotted in red line. It corresponds to the scenario that initial velocity is $v_0=5$ m/s for all vehicles, and bumper to bumper distance $d_0=8$ m.\label{fig:vel5gap8}}
\end{figure*}

We assume surrounding vehicles will ignore other vehicle's turning signals and lane changing behavior unless it is going to collide. They decelerate only when it is necessary to prevent accidents, otherwise they follow leading vehicles closely. 

With our designed planner, the ego vehicle can change lanes successfully, with initial speed $v_0 \in [0.5, 5]$ meters per second and bumper to bumper gap $d_0 \in [4, 10]$ meters. The lane changing process can be found in the  \href{https://drive.google.com/drive/folders/1Z7_Iuaqdd5vBOvA018Uoj8Nv7GQZntHA?usp=sharing}{video recordings}.

We plot relative positions of all vehicles at different times for the case with $v_0=2$ meters per second and $d_0=10$ meters in figure~\ref{fig:vel2gap10}. Black rectangles represent surrounding vehicles and the red rectangle is the ego vehicle, its planning trajectories are marked with red lines. The ego vehicle is seeking the lane changing opportunity in first two seconds, and makes slight lateral movement at the same time. Then it accelerates a little bit and intends to insert into the front right gap between vehicle $2$ and vehicle $3$. By interacting with vehicle $3$, the ego vehicle already has a favorable position when $t=4$ seconds, and completes lane changing at $t=6$ seconds.

The lane changing process with $v_0=5$ meters per second and $d_0=8$ meters is shown in figure~\ref{fig:vel2gap10}. Different from the case in figure~\ref{fig:vel5gap8}, a higher initial speed and smaller gap make it even harder to change lanes. The ego vehicle decelerates a little bit in first two seconds, and then finds a possible trajectory at $t=3$ seconds. By continuing interacting with vehicle $4$, the rear right gap between vehicle $3$ and vehicle $4$ is larger, finally it completes lane changing at $t=5$ seconds.

\begin{figure}[tbp]
    \centering\includegraphics[scale=0.45]{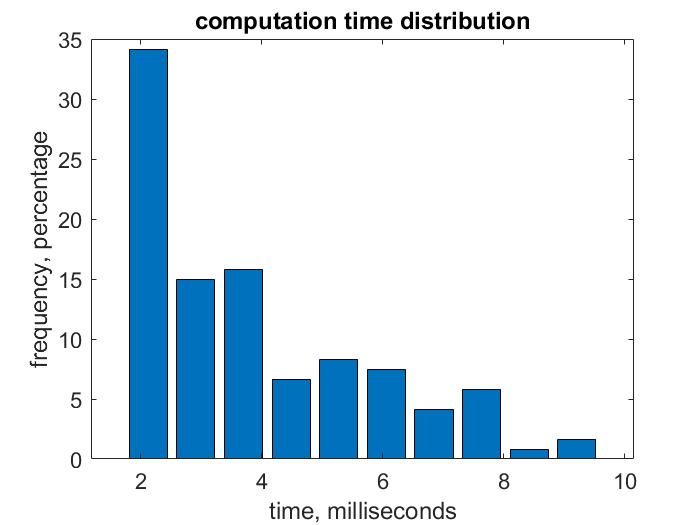}
    \caption{Computation time distribution for the two scenarios corresponding to figure~\ref{fig:vel2gap10} and~\ref{fig:vel5gap8}.\label{fig:compu_time}}
\end{figure}

As the trajectory is planned and updated every $\delta t=0.1$ seconds, we record the computation time every time for the above two scenarios, as shown in figure~\ref{fig:compu_time}. For a laptop with 1.8GHz Intel Core i7-10610U, it costs about 2 milliseconds typically, and less than 10 milliseconds in most cases. The computation efficiency should be able to support real-time planning.

\begin{figure*}[tbp]
    \centering\includegraphics[scale=0.40]{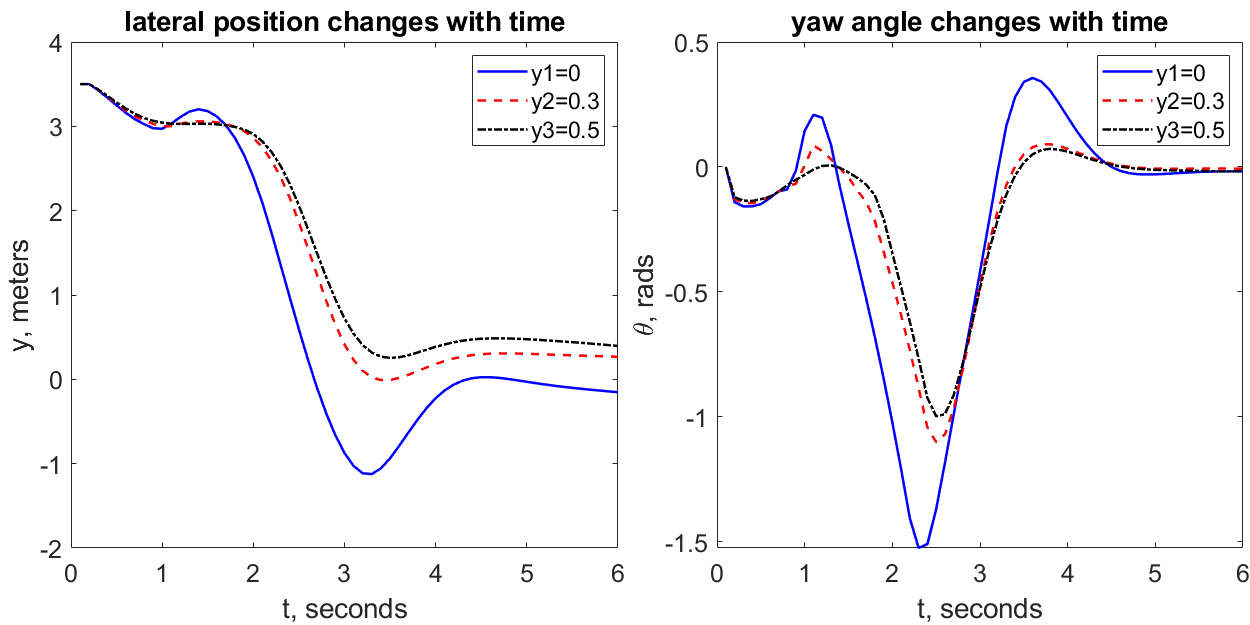}
    \caption{The left and right subplots show the lateral position and yaw angle changes with time, respectively. Blue solid line, red dash line and black dot dash line represent the trajectories corresponding to different desired paths $y1=0$, $y2=0.3$ and $y3=0.5$, respectively.\label{fig:desired_path}}
\end{figure*}

Another observation is that the desired path can considerably influence the smoothness of trajectory, or even determine whether it is doable. It is noted that the center line of the target lane is $y=0$ meters and we compare the ego vehicle's trajectories when the desired path is $y1=0$, $y2=0.3$ and $y3=0.5$ meters in figure~\ref{fig:desired_path}. All these three options for desired path can lead the ego vehicle to change lanes successfully. However, $y1=0$ results in the most lateral position overshooting and large yaw angle, while $y2=0.3$ and $y3=0.5$ can provide a smoother trajectory. To adaptively adjust the desired path in real time may have better performance, e.g., gradually reducing the lateral position of desired path from 0.5 to 0. It can be an interesting extension of this work, which we leave for future work.

%\subsection{Two-lane Highway}

%\subsection{Routing}

\section{Conclusion}\label{sec:conclusion}
In this paper, we propose an interactive planner for lane changing in dense and non-cooperative traffic. Surrounding vehicles will ignore others' turning signals and will decelerate only for collision avoidance. By considering vehicular interactions and leveraging the receding horizon planning technique, our planner can provide safe and smooth trajectories. We implement CILQR algorithm in C++ for solving the optimization problem in real time. In our simulations with eight surrounding vehicles, the performance of our planner is demonstrated and the computation time at each step is less than 10 milliseconds in most cases on a laptop with 1.8GHz Intel Core i7-10610U.

%\section*{Acknowledgment}
%We gratefully acknowledge the support from NSF grants CNS-1839511, CNS-1834701 and UC Lab fees grant LFR-18-548554.

\bibliographystyle{IEEEtran}
\bibliography{Refs}

\end{document}